\documentclass[10pt,journal,compsoc]{IEEEtran}

\RequirePackage[OT1]{fontenc}
\RequirePackage{amsthm,amsmath,amsfonts,amssymb}
\RequirePackage[numbers]{natbib}
\RequirePackage[colorlinks, citecolor=blue, urlcolor=blue]{hyperref}
\RequirePackage{graphicx}

\usepackage{algpseudocode}
\usepackage{algorithm}
\usepackage{url}
\usepackage{subfigure}
\usepackage{ragged2e}
\usepackage[normalem]{ulem}

\graphicspath{{./Images/}} %Path for images

\usepackage{color}
\definecolor{myblue}{RGB}{12, 12, 158}
\definecolor{myorange}{RGB}{245, 150, 12}
\definecolor{mygray}{RGB}{150, 150, 150}
\definecolor{myred}{RGB}{158, 19, 22}
\definecolor{myred2}{RGB}{208, 19, 22}
\definecolor{mygreen}{RGB}{26, 148, 49}
\definecolor{Purple}{RGB}{75, 0, 130}

\newcommand{\given}[1][]{\,#1\lvert\,}

\IEEEoverridecommandlockouts

\usepackage{tikz}
\usepackage{textcomp}
\usepackage{hyperref}
\usepackage{lipsum}

\newcommand\copyrighttext{%
  \footnotesize \textcopyright 2021 IEEE. Personal use of this material is permitted.
  Permission from IEEE must be obtained for all other uses, in any current or future 
  media, including reprinting/republishing this material for advertising or promotional 
  purposes, creating new collective works, for resale or redistribution to servers or 
  lists, or reuse of any copyrighted component of this work in other works. 
  DOI: \href{https://ieeexplore.ieee.org/document/9610018}{10.1109/TPAMI.2021.3124973}}
\newcommand\copyrightnotice{%
\begin{tikzpicture}[remember picture,overlay]
\node[anchor=south,yshift=6pt] at (current page.south) {\fbox{\parbox{\dimexpr\textwidth-\fboxsep-\fboxrule\relax}{\copyrighttext}}};
\end{tikzpicture}%
}

\begin{document}

\title{Regularization of Mixture Models for Robust Principal Graph Learning}

\author{Tony~Bonnaire,~\IEEEmembership{}%
        Aurélien~Decelle~\IEEEmembership{}%
        and~Nabila~Aghanim~\IEEEmembership{}% <-this % stops a space
\IEEEcompsocitemizethanks{\IEEEcompsocthanksitem T. Bonnaire and N. Aghanim are with Université Paris-Saclay, CNRS, Institut d'astrophysique spatiale, 91405, Orsay, France.\protect\\
\IEEEcompsocthanksitem T. Bonnaire and A. Decelle are with Université Paris-Saclay, CNRS, Laboratoire de recherche en informatique, 91190, Gif-sur-Yvette, France. \protect\\
\IEEEcompsocthanksitem A. Decelle is with Departamento de Física Téorica I, Universidad Complutense, 28040 Madrid, Spain}

\thanks{}}

\IEEEtitleabstractindextext{%
\justify
\begin{abstract}
A regularized version of Mixture Models is proposed to learn a principal graph from a distribution of $D$-dimensional datapoints. In the particular case of manifold learning for ridge detection, we assume that the underlying structure can be modeled as a graph acting like a topological prior for the Gaussian clusters turning the problem into a maximum \textit{a posteriori} estimation. Parameters of the model are iteratively estimated through an Expectation-Maximization procedure making the learning of the structure computationally efficient with guaranteed convergence for any graph prior in a polynomial time. We also embed in the formalism a natural way to make the algorithm robust to outliers of the pattern and heteroscedasticity of the manifold sampling coherently with the graph structure. The method uses a graph prior given by the minimum spanning tree that we extend using random sub-samplings of the dataset to take into account cycles that can be observed in the spatial distribution.
\end{abstract}

\begin{IEEEkeywords}
Gaussian Mixture Models, Expectation-Maximization, Graph regularization, Principal Graph, Manifold Learning.
\end{IEEEkeywords}
}

\maketitle

\copyrightnotice
% \IEEEpubidadjcol
%=======================================================
\section{Introduction and contributions} \label{section_introduction}

\IEEEPARstart{D}{ata} often come as a spatially organized set of discrete $D$-dimensional datapoints $\boldsymbol{X} = \{\boldsymbol{x}_i\}_{i=0, \ldots, N}$ with $\boldsymbol{x}_i \in \mathcal{D} \subset \mathbb{R}^D$ sampled from an unknown probability distribution $p$.
These datapoints are usually not spreading uniformly over the entire $\mathbb{R}^D$ space but often result from the sampling of a lower dimensional manifold whose topology and characteristics are linked with the process that generated the data.
Capturing this information, may it be for visual, geometrical or topological analyses of the dataset requires the application of non-linear methods that are parts of the rich field of manifold learning \citep{Roweis2000, Belkin2003}. Revealing and extracting patterns in such datasets with the least prior knowledge is the core of unsupervised machine learning and is usually essential to understand the underlying hidden structure, build models and make predictions.
In many applications, data even appear as standing on a one-dimensional structure as, for instance, GPS measurements collected by vehicles standing on the road network \citep{Ahmed2015}, large-scale galaxy distribution describing the filamentary structure of the cosmic web \citep{Libeskind2017} or vessel networks transporting blood through the human body \citep{Moccia2018}.

The problem of estimating a one-dimensional manifold approximating the underlying distribution of $\boldsymbol{X}$ is a particular case of dimensionality reduction \citep{VanDerMaaten2009} also known as \textit{ridge detection} or \textit{principal curve extraction}. The seminal work of \cite{Hastie1989} provides an intuitive definition of a principal curve as the smooth self-consistent and non-intersecting curve with finite length passing in ``the middle'' of the point cloud distribution. As such, principal curves are a generalization of principal components and relax the straight line condition to describe non-linear relations between variables of multidimensional datasets. Since then, several works followed to obtain an uniform convergence bound \citep{Kegl2000}, give more suitable definitions of these lines as ridges of a probability density \citep[e.g.][]{Tibshirani1992, Ozertem2011} and extend them to higher dimensions (principal surfaces or volumes).

To allow the description of more flexible structures than those imposed by curves and to bypass their inability to represent self-intersecting or cycling topologies, a formulation relying on graph theory to model the one-dimensional structure was introduced in \cite{Gorban2005} and extended in \cite{Gorban2009}. This latter is based on predefined rules for growing the graph and includes regularization terms to limit its overall complexity. Alternatively, \cite{Mao2015} proposes a double optimization scheme with guaranteed convergence for the learning of a tree structure.
In this landscape of methods, only a few address the problem of estimating a principal graph with a proper handling of outliers, which considerably complicates the learning of the graph structure. A built-in robustness to outliers is proposed in \cite{Gorban2016} and \cite{Albergante2020} by discarding from the update of a node position all datapoints beyond a robustness radius $R_0$. However, the choice of $R_0$ is not trivial, scale-dependent and require a careful tuning to deal with uniform background noise \citep{Albergante2020}.

In this work, we use a mixture model to approximate the data distribution $p$ and regularize it over a graph structure to constrain Gaussian centroids to pave this approximation of the underlying one-dimensional structure. The proposed method\footnote{Available at \url{https://git.ias.u-psud.fr/tbonnair/t-rex}.} extends the original presentation of \citep{Tibshirani1992}, that makes use of smooth differentiable curves, to a more general representation of the dataset given by a graph structure. Examples of ridge extractors based on the Subspace Constrained Mean-Shift algorithm \citep[SCMS,][]{Ozertem2011, Genovese2014} offer a mathematically well-defined framework to extract the set of modes and ridges from a kernel density estimate of $p$ using a modified version of the mean-shift algorithm \citep{Fukunaga1975, Cheng1995}. Yet, after convergence of the projected points, it is not trivial to link them to obtain a continuous structure. In the proposed model, this difficulty is overcome by the embedded prior graph structure.

The mixture model formulation offers a natural handling of heteroscedastic patterns through the learning of the local variances of graph nodes, coherently with the local spatial extent of the data. This provides a way to take into account the multi-scale features appearing in many datasets. In contrast, the fixed-scale optimizations proposed by the SCMS or the SimplePPT \cite{Mao2015} algorithms do not permit the description of the local width of the extracted features, even though several scales can be combined. The second advantage of the proposed probabilistic framework is an explicit distribution for outliers, freeing the algorithm from pre-processing steps as required by previous implementations when applied to real-world data \citep{Stanford2000, Chen2014}. We propose here to bypass this difficulty by directly handling outliers in the mixture model through an added uniform background component, making the extraction of the pattern robust to undesirable points with no prior threshold nor processing. It is also more flexible than the fixed trimming radius proposed in \citep{Gorban2016, Albergante2020}.

The overall optimization is based on the Expectation-Maximization \citep[EM,][]{Dempster1977} procedure, allowing us to propose a computationally efficient algorithm for the learning of a principal graph with guaranteed convergence for any prior graph construction.
The presented formulation goes beyond the principal graph construction of \citep{Mao2015, Mao2017}. In particular, the same update equation for node positions can be derived by our formalism in the case of a fixed-variance, the absence of background noise and with a minimum spanning tree prior.

Finally, we introduce a graph construction able to take into account cycling underlying topologies. By exploiting the instability of the minimum spanning tree (MST) to random sub-samplings, we build an average graph consisting in the union of edges from the MST and of high probability edges. We show in several applications that this graph embeds cycles representing one-dimensional holes in the dataset independently of their scales or sizes and without giving them a formal definition. This construction is compared to what is proposed by the framework of persistent homology \citep{Kurlin2015}. Even though mathematically well-defined, this framework does not allow the easy extraction of relevant edges independently of their size and create spurious edges irrelevant for the underlying pattern that are not captured in the proposed empirical construction.

% Organization of the paper
The paper is organized such that Section \ref{section_background} sets notations and provides technical information about mixture models, Expectation-Maximization algorithm and graph theory.  Section \ref{section_method} presents the regularized mixture model embedding the multiple prior distributions, gives explicit equations for parameter estimation, details the algorithm and reviews some of its technical aspects. It also describes the graph construction based on the average of minimum spanning trees. Finally, Section \ref{section_applications} provides and discuss some direct applications of the algorithm on real-life problematics in which the proposed features are relevant.

%==========================================================================

\section{Technical background}  \label{section_background}

\subsection{Mixture models and Expectation-Maximization algorithm}

Observed data are usually not directly drawn from known probability distributions but come with multiple modes and complex shapes.
Mixture models tackle this complexity through a linear combination of $K$ known laws with unknown parameters by writing the probability of a datapoint $\boldsymbol{x}$ as
\begin{equation} \label{MM}
    p(\boldsymbol{x} \given \boldsymbol{\Theta}) = \sum_{k=1}^{K} \pi_k f(\boldsymbol{x}, \boldsymbol{\theta}_k),
\end{equation}
with $\boldsymbol{\Theta} = \{\pi_1, \ldots, \pi_K, \boldsymbol{\theta}_1, \dots, \boldsymbol{\theta}_K\}$ the set of model parameters, $\pi_k \geq 0$ the mixing coefficient of component $k$ such that $\sum_{k=1}^K \pi_k = 1$ and $f(\boldsymbol{x}, \boldsymbol{\theta}_k)$ the probability distribution describing the $k^\text{th}$ component. The particular Gaussian case of mixture models can be obtained by considering $f(\boldsymbol{x}, \boldsymbol{\theta}_k) = \mathcal{N}(\boldsymbol{x}, \boldsymbol{\theta}_k)$, where $\mathcal{N}(\boldsymbol{x}, \boldsymbol{\theta}_k)$ is the Gaussian distribution centered on $\boldsymbol{\mu}_k$ and with variance $\Sigma_k$. Thanks to their ability to easily model complex data and their tractability, Gaussian mixture models (GMMs) are the most used mixture distributions in the machine learning literature, mainly as a clustering algorithm to separate a dataset into $K$ components.

What makes mixture models so used in practice is the EM algorithm \citep{Dempster1977}, allowing the maximization of the log-likelihood in a polynomial time.
Assuming an independent and identically distributed set of datapoints $\boldsymbol{X} \in \mathbb{R}^{N\times D}$, the log-likelihood of the model defined by Eq.~\eqref{MM} can be written
\begin{equation} \label{eq:log_likelihood}
    \mathcal{L}(\boldsymbol{\Theta}; \boldsymbol{X}) = \sum_{i=1}^{N} \log \left( \sum_{k=1}^K \pi_k f(\boldsymbol{x}_i, \boldsymbol{\theta}_k) \right),
\end{equation}
which cannot be analytically optimized due to the summation inside the logarithm function.
To counter that difficulty, the idea behind EM is to optimize iteratively a simpler function acting like a lower-bound on this log-likelihood. Introducing a set of latent variables $\boldsymbol{Z} = \{z_i\}_{i=1}^N$ encoding the partition of the dataset such that $z_i \in \{1, \ldots, K\}$ denotes which of the $K$ Gaussian components $x_i$ belongs to, the log-likelihood conditioned on the value of $\boldsymbol{Z}$ can hence be expressed as
\begin{equation}  \label{complete_logL}
    \mathcal{L}(\boldsymbol{\Theta}; \boldsymbol{X}, \boldsymbol{Z}) =  \sum_{i=1}^N \log \left(\pi_{z_i} \, f(\boldsymbol{x}_i, \boldsymbol{\theta}_{z_i}) \right).
\end{equation}

EM states that \eqref{complete_logL} can be maximized through an iterative procedure involving two alternating steps. Assuming we stand at the iteration $t$, the E-step uses the current values of the parameters $\boldsymbol{\Theta}^{(t)}$ to estimate the posterior distribution of the latent variables $p(\boldsymbol{Z} \given \boldsymbol{X}, \boldsymbol{\Theta}^{(t)}$) that will allow the computation of the expectation over the completed log-likelihood
\begin{align} \label{Estep}
    Q(\boldsymbol{\Theta}, \boldsymbol{\Theta}^{(t)}) &= \mathbb{E}_{\boldsymbol{Z}\given\boldsymbol{X}, \boldsymbol{\Theta}^{(t)}}\left\{\mathcal{L}(\boldsymbol{\Theta}; \boldsymbol{X}, \boldsymbol{Z}) \right\}
\end{align}
The M-step then updates parameters by maximizing \eqref{Estep} as
\begin{equation} \label{Mstep}
    \boldsymbol{\Theta}^{(t+1)} = \operatorname*{argmax}_{\boldsymbol{\Theta}} \, Q(\boldsymbol{\Theta}, \boldsymbol{\Theta}^{(t)}).
\end{equation}
By alternating both Expectation and Maximization steps, EM guarantees the convergence toward a local maximum of the log-likelihood \eqref{eq:log_likelihood}.

\subsection{Graph theory}

Let us consider a graph $\mathcal{G}=(\mathcal{V}, \mathcal{E}, \mathcal{W})$ as an ensemble of vertices $\mathcal{V}$, edges $\mathcal{E} = \{(i,j) | i\sim j\}$ where $i\sim j$ denotes that nodes indices $i$ and $j$ are linked together and edge weights $\mathcal{W} = \{w_{ij}\}_{(i,j)\in\mathcal{V}^2}$ such that $\forall (i,j)\in\mathcal{V}^2, w_{ij} \geq 0$ being non-zero when $i$ and $j$ are linked. In this work, all considered graphs are undirected meaning that $\mathcal{E}$ is an unordered set. To further use graphs in linear algebra contexts, we rely on matrix representations given by the adjacency matrix $\boldsymbol{A}$, a symmetric $\rvert{\mathcal{V}\lvert}\times\rvert{\mathcal{V}\lvert}$ matrix encoding whether two vertices are linked or not, i.e. takes value $1$ at position $(i,j)$ if $i\sim j$ and 0 otherwise.

In this work, we assume that a spatial graph $\mathcal{G}$ embedded in $\mathbb{R}^D$ with $K$ vertices at positions $\{\boldsymbol{\mu}_k\}_{k=1}^K$ can be computed to geometrically approximate the data space.
In particular, we use $\lVert \boldsymbol{\mu} \rVert^2_\mathcal{G}$ with $\boldsymbol{\mu} = (\boldsymbol{\mu}_1, \ldots, \boldsymbol{\mu}_k)^\text{T}$ as a measure of the smoothness of the graph structure \citep{Smola2003} to constrain the mixture model. Mathematically, this quantity is similar to the usually used $L_2$-norm to achieve that purpose, but computed over the non-Euclidean structure of the graph,
\begin{align} \label{Laplace}
        \lVert \boldsymbol{\mu} \rVert^2_\mathcal{G} &= \sum_{i=1}^K \sum_{j=1}^K a_{ij} \rVert \boldsymbol{\mu}_i - \boldsymbol{\mu}_j \Vert^2_2 \nonumber \\
        &= 2\, \text{Tr} \{ \boldsymbol{\mu^\text{T} L \mu} \},
\end{align}
where $\boldsymbol{L} = \boldsymbol{D} - \boldsymbol{A}$ is the Laplacian matrix, symmetric in the undirected case, with $\boldsymbol{D}$ a diagonal $K\times K$ matrix such that $d_{kk} = \sum_{j=1}^K a_{kj}$ is the degree of vertex $k$.

Usually, for a given dataset, the underlying graph modeling of the manifold is not known and not obvious to build. There are many different kind of graphs that can be used to approximate the discrete data structure based on topological priors one has at hand. In this work, we particularly focus on two kinds of structures with different topologies to model the data: the MST and an extension of it allowing the representation of cycles in the structure that we present in Sect. \ref{subsection_averageGraph}.

The MST of a collection of $K$ vertices corresponds to the set of $K-1$ edges with a tree topology (assumes no self loop and the existence of a unique path between any pair of nodes) minimizing the total sum of edge weights to reach all vertices. We can formulate it as an integer programming problem
\begin{equation} \label{MST_opt_problem}
\boldsymbol{A}_{\text{MST}} = \operatorname*{argmin}_{\boldsymbol{A} \in \mathcal{A}} \sum_{i=1}^K\sum_{j=1}^K a_{ij} \lVert \boldsymbol{\mu}_i - \boldsymbol{\mu}_j \rVert^2_2
\end{equation}
where $\mathcal{A}$ is the set of matrices such that $\boldsymbol{A} = \boldsymbol{A}^\text{T}$, $a_{ij} \in \{0,1\}$, $\sum_{i=1}^K \sum_{j=1}^K a_{ij} = 2\left( K-1 \right)$, $\forall k \in \{1, \ldots, K\}, a_{kk} = 0$ and $\forall S \subseteq \mathcal{V}, \sum_{i\in S} \sum_{j\in S} a_{ij} \leq \lvert S \rvert - 1$. The first two constraints encode the undirected graph definition, the third one denotes a graph with $K-1$ edges, the fourth the absence of self loops in the structure and the last one requires that any subset of $\rvert \mathcal{S} \lvert$ vertices has at most $\rvert \mathcal{S} \lvert - 1$ edges. By relaxing the integer constraint to $a_{ij} > 0$, the problem can be solved in quasi-linear time with the number of edges using Kruskal's algorithm \citep{Kruskal1956}.

%==========================================================================
\section{Method}  \label{section_method}

\subsection{Regularized mixture model} \label{section_RMM}

In what follows, we focus on the particular case of GMMs hence assuming that the sampling noise around the ridge is Gaussian. From the principal curve point of view, these Gaussian clusters are the projected points standing in the middle of the data distribution. One of the drawbacks of usual mean square projection definitions of principal curves is their sensitivity to outliers, tending to bias the curve because all datapoints have the same weights. Real-world datasets, however, often come with outliers or noisy measurements that are not part of the underlying pattern that generated the data we aim to extract.
In the mixture model context, similarly to what is done in \citep{Stanford2000} for clustering purposes, we tackle this problem by assuming an explicit form for the distribution of outliers. To make the representation robust to these noisy measurements, an additional uniform background distribution is incorporated leading the mixture model to
\begin{equation} \label{model}
    p(\boldsymbol{x} \given \boldsymbol{\Theta}) = \sum_{k=1}^{K} \pi_k \, \mathcal{N}(\boldsymbol{x}, \boldsymbol{\theta}_k) + \alpha \rho(\boldsymbol{x} ),
\end{equation}
where $\sum_{k=1}^K \pi_k \, + \, \alpha = 1$ and $\rho(\boldsymbol{x}) = \left[ \int_{\mathbb{R}^D} 1_\mathcal{D}(\boldsymbol{x}) \, \text{d}\boldsymbol{x} \right]^{-1}$ the uniform background distribution with $\mathcal{D}$ the support of the dataset. $\alpha$ being a parameter of the model estimated during the learning, the sole prior information required by the framework about outliers is $\rho(\boldsymbol{x})$, a constant estimating the inverse of the data support volume. In the rest of the paper, we rely on the estimate of the volume of the convex hull of the $N$ datapoints. The uniform background distribution for outliers is not an unique choice and could be replaced by any other fitting a prior knowledge. In the absence of external information about a positional dependency for outliers, the uniform distribution is however a natural choice in the sense it maximizes the entropy over the closed domain $\mathcal{D}$.

Because $D$-dimensional datasets often lie on a lower dimensional support $d < D$, we can include this knowledge into the model as a prior constraining the parameter space $\boldsymbol{\Theta}$.
In the context of ridge extraction, $d=1$ and, keeping the underlying topology as general as possible, it is assumed that a graph $\mathcal{G}$ can approximate this one-dimensional structure from which $\boldsymbol{X}$ is drawn. The topological prior $\mathcal{G}$ can be embedded as a penalty term on the log-likelihood adding a smoothness condition on the set of centers $\boldsymbol{\mu}$ through $\lVert \boldsymbol{\mu}\rVert^2_\mathcal{G}$ defined in Eq.~\eqref{Laplace}. Formally, we write
\begin{equation} \label{prior_mu}
    \log p(\boldsymbol{\mu}) = -\frac{\lambda_{\mu}}{2} \lVert \boldsymbol{\mu}\rVert^2_\mathcal{G},
\end{equation}
where $\lambda_{\mu} \geq 0$ is a parameter controlling the force of the penalty that we later refer to as ``regularization parameter''.
This constraint is very similar to the one introduced in \cite{Tibshirani1992} where the curvature measure of the differentiable curve is used instead to regularize the log-likelihood. Here, the represented structure and its topology are made more general by using a graph $\mathcal{G}$. The same type of constraints as in \cite{Yuille1990, Tibshirani1992} can be obtained by restricting the graph to a chain topology with elements of the Laplacian matrix $l_{ij} = 2\delta_{i,i} - \delta_{i,j+1} - \delta_{i,j-1}$, where $\delta_{i,j}$ denotes the Kronecker delta function. The non-Euclidean nature of the approximated manifold is encoded in the Laplacian matrix $\boldsymbol{L}$ and $\lVert \boldsymbol{\mu}\rVert^2_\mathcal{G}$ hence plays the role of $\int_\mathcal{G} \lVert \nabla \boldsymbol{\mu}_k \rVert^2$, a measure of the manifold smoothness \cite{Belkin2001} when approximated by $\mathcal{G}$. From a pure Bayesian view, Eq.~\eqref{prior_mu} also corresponds to a Gaussian prior on the edge weights distribution with a precision $\lambda_\mu$.

Natural manifestations of heteroscedastic patterns, such as trees or blood vessels, exhibit a linear evolution of the sampling size without sudden variations. For instance, the width of a tree is continuously expanding from small branches to the trunk and the size of the vessel network is smoothly enlarging from capillaries to wide arteries. For these physically motivated reasons, the local size of the underlying continuous structure as measured by variances of the model can be considered to evolve smoothly along the graph. To incorporate this idea in the formalism, we use again the non-Euclidean proximity measure on the graph structure for variances update.
To this end, an additional prior distribution on variances is used, based on the local neighborhood of a node.
In what follows, we pave the point-cloud distribution with spherical Gaussian clusters such that $\forall k \in \{1, \ldots, K \}, \boldsymbol{\Sigma}_k = \sigma_k^2 \, \boldsymbol{I}_D$. To obtain a closed-form expression of the new update equation, we use the conjugate prior for variances of a Gaussian likelihood. Formally, we rely on the inverse-Gamma distribution with shape parameter $1+\lambda_{\sigma}$ and scale parameter $\lambda_{\sigma} \sigma_{\mathcal{N}_k}^2$ defined such that the mode of the distribution is located at the mean variance of the neighboring nodes, namely $\sigma_{\mathcal{N}_k}^2 = 1/\lvert \mathcal{N}_k \rvert \sum_{i \in \mathcal{N}_k} \sigma_i^2$ with $\mathcal{N}_k = \{i \given a_{ik} = 1\}$ and $\lvert \mathcal{N}_k \rvert = d_{kk}$ the degree of node $k$. Mathematically, we write
\begin{equation} \label{prior_sigma}
    \log p(\sigma_k^2) = -\lambda_{\sigma} \left[ \log \sigma_k^2 + \sigma_{\mathcal{N}_k}^2/\sigma_k^2 \right] + \text{cst},
\end{equation}
the prior distribution for the $\sigma_k^2$ parameter and use the same $\lambda_{\sigma} \geq 0$ to constrain all components.

Finally, a prior distribution is added for mixing coefficients to avoid singular solutions to happen in situations in which nodes of the graph are paving almost empty regions leading to $\pi_k \to 0$. This can be achieved by assuming a Gaussian prior centered on uniform coefficients on the set of datapoints not represented by the background noise, namely $\left(1-\alpha\right)/K$. Hence, we have
\begin{equation} \label{prior_weights}
    \log p(\pi_k) = -\frac{\lambda_{\pi}}{2} \left[\frac{1 - \alpha}{K} - \pi_k\right]^2 + \text{cst},
\end{equation}
where $\lambda_{\pi} \geq 0$ controls the force of this prior.

The full prior distribution on the parameter set $\boldsymbol{\Theta}$ can hence be written as the summation of all individual terms,
\begin{equation} \label{prior}
    \log p(\boldsymbol{\Theta}) = \log p(\boldsymbol{\mu}) + \sum_{k=1}^K \log p(\sigma_k^2) + \sum_{k=1}^K \log p(\pi_k).
\end{equation}

Optimal values of the parameters are now obtained by maximizing the posterior distribution defined as $\log p(\boldsymbol{\Theta} \given \boldsymbol{x}) \propto \log p(\boldsymbol{x} \given \boldsymbol{\Theta}) + \log p(\boldsymbol{\Theta})$ turning the problem into a maximum \textit{a posteriori} estimation of the parameters for which the EM algorithm can still be used.

As the additional term only depends on $\boldsymbol{\Theta}$, the E-step remains unchanged while the M-step now seeks to solve
\begin{equation} \label{regularized_mstep}
    \boldsymbol{\Theta}^{(t+1)} = \operatorname*{argmax}_{\boldsymbol{\Theta}} \, Q(\boldsymbol{\Theta}, \boldsymbol{\Theta}^{(t)}) + \log p(\boldsymbol{\Theta}).
\end{equation}

For the log-likelihood as defined by the particular mixture model \eqref{model} and the prior defined by Eq.~\eqref{prior}, responsibilities of Gaussian and uniform background components, respectively noted $p_{ik}$ and $p_i^{\text{bkg}}$, can be computed during the E-step as $p(\boldsymbol{Z} \given \boldsymbol{X}, \boldsymbol{\Theta}^{(t)})$ using Bayes' theorem and the current parameter values $\boldsymbol{\Theta}^{(t)}$
\begin{equation}  \label{Eupdate}
    \left\{
    \renewcommand*{\arraystretch}{3.0}
    \begin{array}{l}
        \displaystyle p_{i}^\text{bkg} = \frac{\alpha \rho(\boldsymbol{x}_i)}{\sum_{j=1}^K \pi_j \, \mathcal{N}(\boldsymbol{x}_i \given \boldsymbol{\theta}_j) + \alpha \rho(\boldsymbol{x}_i)}, \\
        \displaystyle p_{ik} = \frac{\pi_k \, \mathcal{N}(\boldsymbol{x}_i \given \boldsymbol{\theta}_k)}{\sum_{j=1}^K \pi_j \, \mathcal{N}(\boldsymbol{x}_i \given \boldsymbol{\theta}_j) + \alpha \rho(\boldsymbol{x}_i)}.
    \end{array}
    \right.
\end{equation}
Update equations for each parameter are then derived during the M-step of Eq.~\eqref{regularized_mstep} as
\begin{equation}   \label{Mupdate}
    \left\{
    \renewcommand*{\arraystretch}{3.0}
    \begin{array}{l}
        \alpha^{(t+1)} = \displaystyle\frac{1}{N} \sum_{i=1}^N p_{i}^{\text{bkg}}, \\
        \pi_k^{(t+1)} = \displaystyle \frac{1/N\sum_{i=1}^N p_{ik} + \lambda_{\pi}\left(1 - \alpha^{(t+1)}\right)/K}{1+\lambda_{\pi}}, \\
        \displaystyle \boldsymbol{\mu}_k^{(t+1)} = \frac{\sum_{i=1}^N \boldsymbol{x}_i \, p_{ik}/\sigma_k^2 + 2\lambda_{\mu}\sum_{j=1}^K a_{kj}\boldsymbol{\mu}_j^{(t+1)}}{\sum_{i=1}^N p_{ik}/\sigma_k^2 + 2\lambda_{\mu}\sum_{j=1}^K a_{kj}}, \\
        \sigma_k^{(t+1)} = \displaystyle \left[ \frac{\sum_{i=1}^N p_{ik} \lVert \boldsymbol{x}_i - \boldsymbol{\mu}_k \rVert_2^2 + 4 \lambda_{\sigma} \sigma_{\mathcal{N}_k}^2}{D\sum_{i=1}^N p_{ik} + 4\lambda_{\sigma}} \right]^{1/2}.
    \end{array}
    \right.
\end{equation} 
When not specified, parameters not indexed by time correspond to time $t$. This is especially important in the E-step and in the computation of $\boldsymbol{\mu}_k^{(t+1)}$. It is indeed interesting to note the contribution of all $\boldsymbol{\mu}_j^{(t+1)}$ in the update of $\boldsymbol{\mu}_k^{(t+1)}$ in Eq.~\eqref{Mupdate}. It is hence more convenient to write this update equation matricially for $\boldsymbol{\mu} \in \mathbb{R}^{K\times D}$.
\begin{equation}  \label{eq:matrix_update}
\boldsymbol{\mu}^{(t+1)}= \left[ \boldsymbol{\Gamma} \boldsymbol{S}^{-1} + 2\lambda_{\mu} \boldsymbol{L}\right]^{-1} \boldsymbol{S}^{-1} \boldsymbol{R}^\text{T} \boldsymbol{X},
\end{equation}
where $\boldsymbol{S}$ is a diagonal $K\times K$ matrix with $s_{kk} = \sigma_k^2$, $\boldsymbol{\Gamma}$ a diagonal $K \times K$ matrix of average datapoints explained by the $k^{\text{th}}$ component, i.e. $\gamma_{kk} = \sum_{i=1}^N p_{ik}$ and $\boldsymbol{R} \in \mathbb{R}^{N \times K}$ the responsibility matrix for Gaussian components such that $r_{ik} = p_{ik}$.

%==========================================================================
\subsection{Algorithm and illustration} \label{section_algo}

\begin{algorithm}
    \caption{Graph regularized mixture model}
    \label{Algo:GRMM}
    \begin{algorithmic}
        \Statex \textbf{Input:} Data: $\boldsymbol{X} \in \mathbb{R}^{N\times D}$, hyper-parameters: $\mathcal{Y} = \{\lambda_{\mu}, \lambda_{\sigma}, \lambda_{\pi}\}$, initialization: $\boldsymbol{\Theta}^{(0)}$, $\mathcal{G}^{(0)}$.
        \Statex \textbf{Output:} $\boldsymbol{\Theta}^{(t)}$ and $\mathcal{G}^{(t)}$.
        
        \While{$\mathcal{R}_t \geq \epsilon$}
        
            \State Compute the adjacency matrix $\boldsymbol{A}$ of $\mathcal{G}^{(t)}$
            
            \State \textbf{E-step:} Compute responsibilities $p_{ik}$ and $p_i^\text{bkg}$ using equations \eqref{Eupdate}
            
            \State \textbf{M-step:} Compute new parameters $\boldsymbol{\Theta}_t$ based on responsibilities using equations \eqref{Mupdate} and \eqref{eq:matrix_update}
            
            \State Compute the increment in log posterior $\mathcal{R}_t = \log\,p(\boldsymbol{\Theta}^{(t)} \given \boldsymbol{X}) - \log\,p(\boldsymbol{\Theta}^{(t-1)} \given \boldsymbol{X})$
            
            \State \textbf{Optional:} Update graph topology by recomputing $\mathcal{G}^{(t)}$ on the new set $\{\boldsymbol{\mu}_k\}_{k=1}^K$
            
        \EndWhile
    \end{algorithmic}
\end{algorithm}

Algorithm \eqref{Algo:GRMM} sums up all the steps for the learning of the proposed graph regularized mixture model (GRMM) from a given set of measurements $\boldsymbol{X}$ and a graph prior $\mathcal{G}$. 
It extends the GMM by embedding a topological prior through $\mathcal{G}$ constraining the position of centers, propose an handling of outliers and the learning of the local widths.
In our model, the topology of the graph can be either fixed or updated during the optimization. This is particularly important since the prior approximation on the noisy pattern with outliers may not be suitable to describe the final smooth underlying structure. For instance, a bifurcation in the prior graph may not be relevant in the final one. Consequently, it may be useful to update the topological prior based on the current estimate of $\{\boldsymbol{\mu}\}_{k=1}^K$. When updating the topology at each iteration, $\boldsymbol{A}$ cannot be understood as a prior anymore but has to be included as a parameter in the model making the chosen graph construction itself acting like a prior. In the case of fitting an MST topology, we can embed the integer programming optimization problem definition of the MST from Eq.~\eqref{MST_opt_problem} in the regularized log-likelihood hence leading to estimate both $\boldsymbol{\Theta}$ and $\boldsymbol{A}$ in the M-step which simply involves computing the new graph $\mathcal{G}_t$ at iteration $t$ given a graph construction process and the positions $\{\boldsymbol{\mu}\}_{k=1}^K$.

\begin{figure}
    \centering
    \includegraphics[width=1\linewidth]{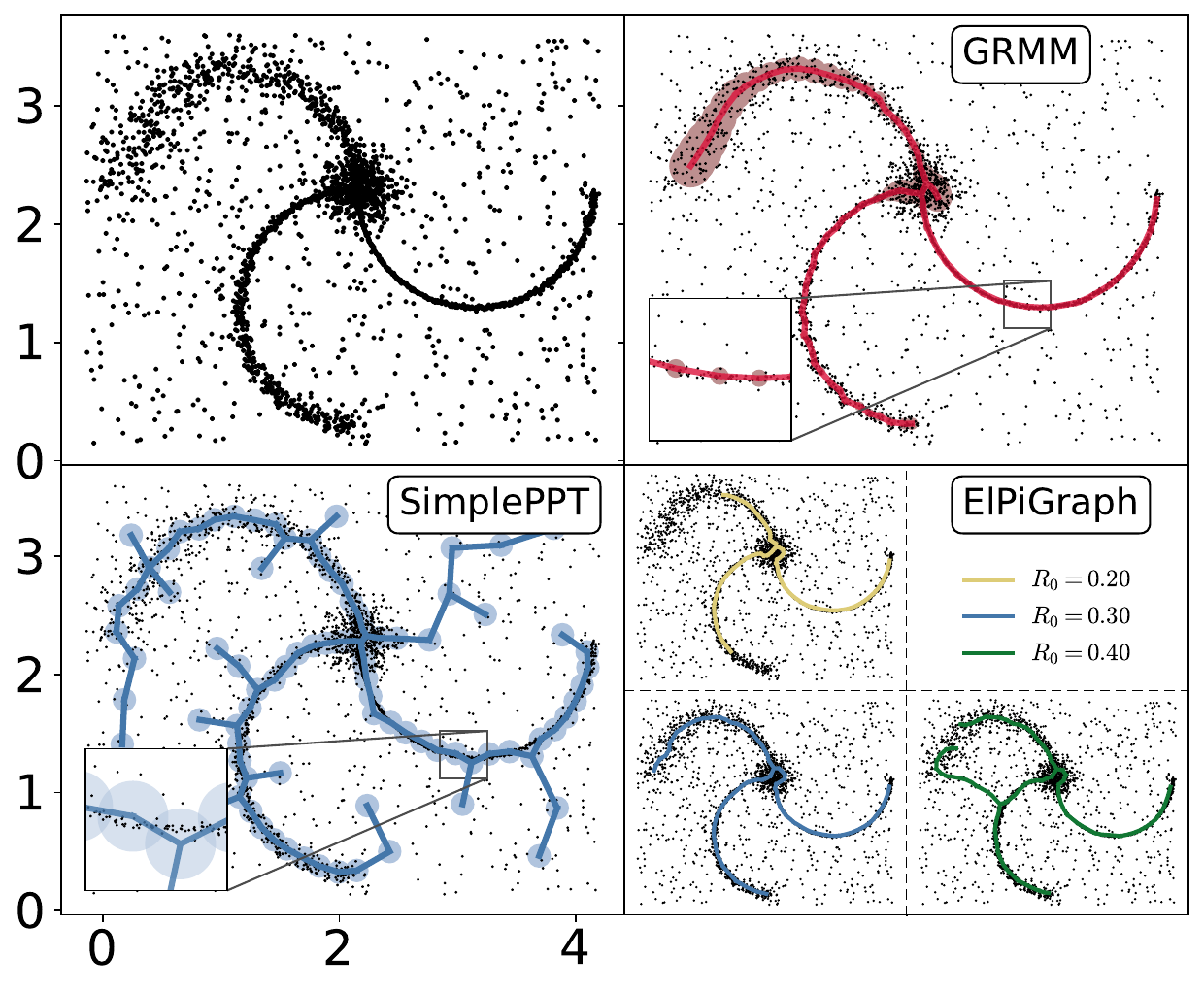}
    
    \caption{Illustrative comparison of three procedures to learn a principal graph on an artificial dataset made of $N=2666$ datapoints with three branches with linearly evolving standard deviations converging to a spherical Gaussian cluster shown in the top left panel.
    Top right is the principal graph learned by the proposed GRMM algorithm \eqref{Algo:GRMM} with $K=100$ initialized randomly over $\boldsymbol{X}$, $\mathcal{Y} = \{5/\sigma_0^2, 10, 1\}$ and $\sigma_0^2=0.01$. Bottom left is the one from the SimplePPT algorithm \citep{Mao2015, Mao2017} with $\sigma = \sigma_0$ and identical initialization nodes. In both cases, the shaded areas correspond to the 1-$\sigma_k$ circles centered on $\boldsymbol{\mu}_k$.
    The bottom right panel is the result from the ElPiGraph procedure \citep{Albergante2020} with $100$ nodes initialized with $70$ taken randomly over $\boldsymbol{X}$ and for different values of the trimming radius $R_0$  with the same elasticity parameters ($\lambda = 0.01$, the length constraint and $\mu=0.01$, the bending constraint).}
    
    \label{fig:Illustration_algo}
\end{figure}

The top right panel of Fig.~\ref{fig:Illustration_algo} shows the graph learned by the GRMM Algorithm \eqref{Algo:GRMM} for a dataset made of three branches with linearly evolving standard deviations $\sigma \in \left[0.015, 0.15\right]$ and $25\%$ background noise uniformly added in the square. The resulting graph visually shows a smoothly evolving adaptive scale along the structure. Even though the graph is initialized with only $K=100$ nodes taken randomly over $\boldsymbol{X}$, there are no final nodes standing outside of the pattern and the resulting estimate of the background level is $25\%$. The bottom left panel of Fig.~\ref{fig:Illustration_algo} illustrates the principal graph learned by the SimplePPT algorithm on the same dataset and using the same parameters and initialization as our construction. Many branches are falling in the background noise and, because of its fixed-variance scheme, it fails at capturing the ridges of both the large and small variance branches. Choosing a large value for $\sigma_0^2$ biases the graph in the small branch, while a low one irremediably creates a lot of spurious branches in the large-variance part of the pattern, even in the absence of outliers. The sensitivity of the graph to outliers is handled by the ElPiGraph framework \citep{Albergante2020} through a trimming radius $R_0$ discarding distant datapoints from the update of a node position. In the bottom right panel of Fig.~\ref{fig:Illustration_algo} can be found three realizations of this algorithm with different values of $R_0$, showing the robustness of ElPiGraph to uniform background noise with a fine-tuned value. The choice of $R_0$ however remains scale-dependent and can alter the recovered structures in case of heteroscedastic patterns. It is also not obvious to tune in real applications and comes in addition to other hyper-parameters already complex to choose. Moreover, the resulting graph do not embed a description of the local width as opposed to the GRMM which fits all parts of the pattern in its center. Note nonetheless that the ElPiGraph algorithm handles better the dense area at the center of the dataset when $R_0 = 0.3$ with a single bifurcation thanks to an added regularization term to the cost function minimizing the overall graph complexity (length as Eq.~\eqref{prior_mu} but also bending).

%Figure: Illustration of the algorithm

\subsection{Average graph prior}  \label{subsection_averageGraph}

Although the MST exhibits some nice features, especially the scale invariance and the absence of free parameter, it has a limited representative power for general datasets due to its tree topology that cannot represent cycles. By using a non-cycling topology, the regularized graph becomes even more inaccurate if the data embeds holes because the optimization shortens graph extremities hence emphasizing the absence of loops.
Several graph constructions allow such a feature, as for instance the $k$-nearest neighbors graph. This solution would however create a lot of redundancy in the structure and long-range edges for isolated nodes standing in outliers.

Because the MST results from a global minimization of the total Euclidean distance as seen in Eq.~\eqref{MST_opt_problem}, the obtained preferential path is very sensitive to random removals of datapoints. We exploit this particular aspect by merging the idea of a graph with the minimum total length and the handling of one-dimensional holes in the dataset to propose an empirical construction based on the computation of MSTs obtained from a set of $B$ realizations of random sub-samplings of the set of nodes. Formally, the average adjacency matrix is $\bar{\boldsymbol{A}} = \sum_{b=1}^B \boldsymbol{A}_b / B$, where $\boldsymbol{A}_b$ is the adjacency matrix of the MST computed the $b^\text{th}$ random sub-sampling of $\boldsymbol{\mu}$ using a fraction $K_b/K$ of points. This approach, proposed in the context of pattern extraction to provide an uncertainty measurement \citep{Chen2014, Bonnaire2020, Albergante2020}, is exploited in the present case to combine all the realizations to obtain a single graph construction. This operation is also done \textit{a priori} on non-regularized MSTs hence reducing the computational cost by avoiding the need of running Algorithm \eqref{Algo:GRMM} multiple times.

\begin{figure}
    \centering
    \subfigure{\includegraphics[width=1\linewidth]{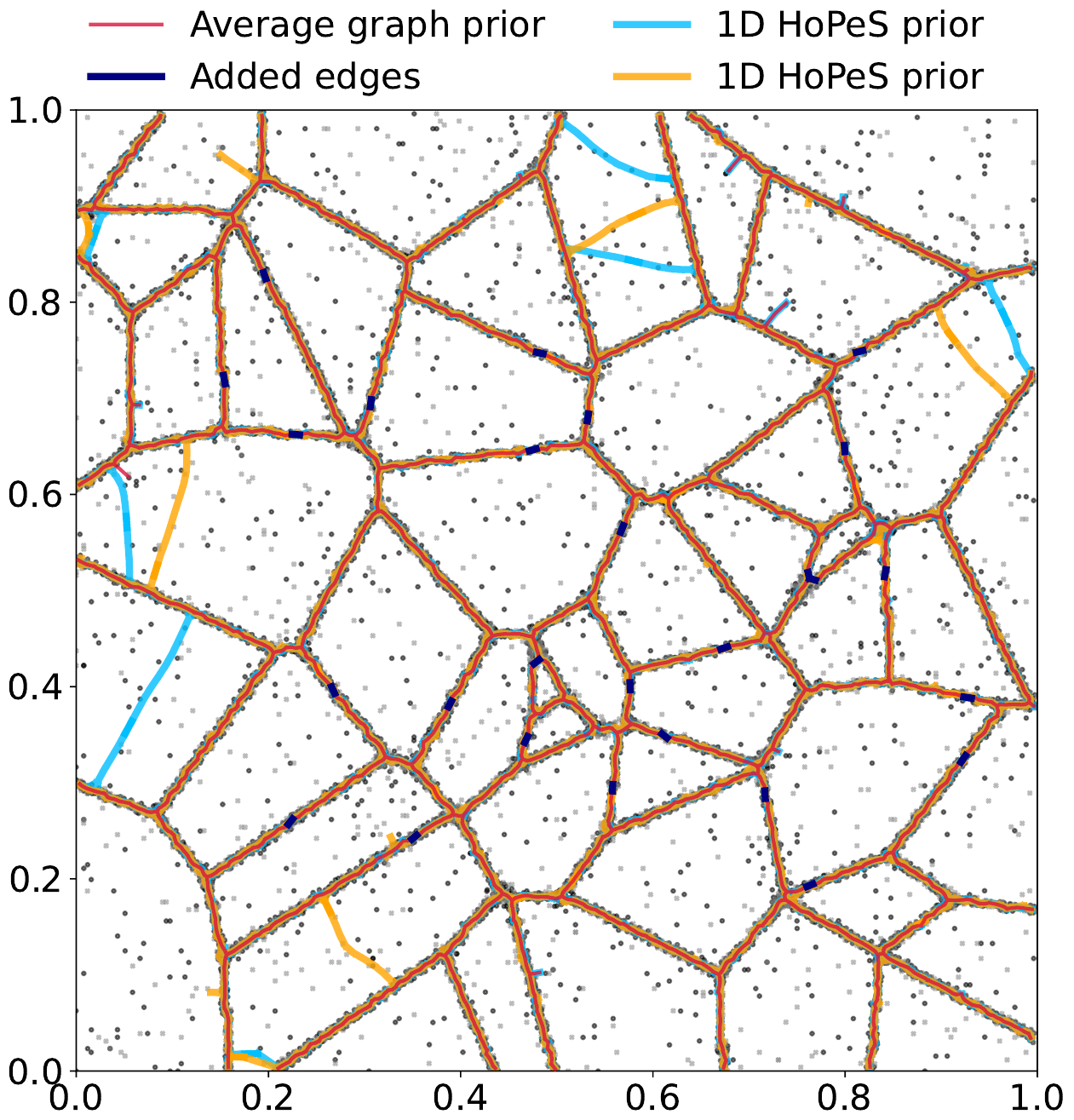}}   %Voronoi_noise_clr_rev_v3
    \subfigure{\includegraphics[width=1\linewidth]{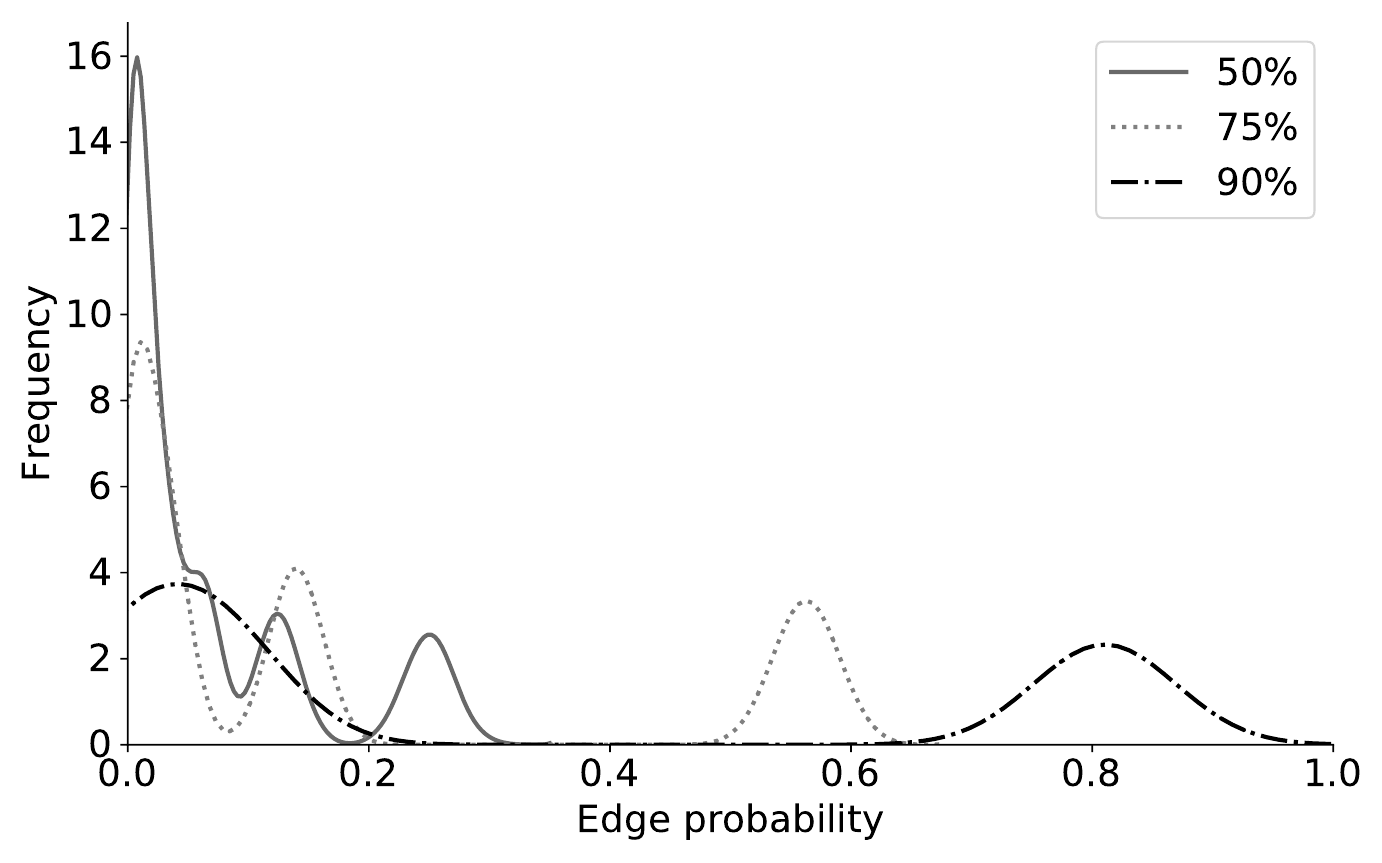}}
    
    \caption{\textit{(top)} Black points and gray crosses are two sampling realizations of artificial 2D datasets obtained from a Voronoi pattern with $N=9249$ points. Red edges are those from the MST over the final regularized structure with $K=3000$ nodes and bold dark blue ones are those added by the high probability mode with $B=500$, $K_b/K = 0.75$, and thresholded at $m=0.35$. The turquoise blue and orange lines are the set of edges from the regularized graph respectively computed from the two noisy realizations with the 1D HoPeS prior \citep{Kurlin2015}. 
    %When not explicitly visible, the three lines overlap perfectly.
    \textit{(bottom)} Probability distributions of edge probabilities $\left(\bar{\boldsymbol{A}}\right)_{ij}$ for several ratios $K_b/K$.}
    \label{fig:Voronoi}
\end{figure}

The averaged adjacency matrix $\bar{\boldsymbol{A}}$ represents the frequency of appearance of an edge during $B$ realizations of MST and thus the probability, for each pair of centers, to be linked. To illustrate the method, we build a discrete Voronoi dataset consisting of intersecting straight lines surrounding low-density areas visible on the top panel of Fig.~\ref{fig:Voronoi}. The bottom panel shows the distribution of edge probabilities for several ratios $K_b/K$. For a large range of values of this ratio, we clearly distinguish two populations of edges: those with high probabilities and those with much lower ones. In the high probability mode, we retrieve, in the three cases, all $K-1$ edges of the original MST computed over $\boldsymbol{\mu}$ plus 27 new ones which corresponds to the exact number of closed cycles in the artificial dataset. These additional edges are shown in bold dark blue in the top panel of Fig.~\ref{fig:Voronoi}. When the ratio $K_b/K$ becomes smaller, the high probability mode tends to be centered at lower and lower probability until the pattern is so much altered by the sub-sampling that it is not retrieved. The high probability mode then becomes centered at lower and lower values. The proposed method hence allows to take advantage of the inherent instability of the MST to recover additional edges inducing cycles independently of their scale. It solely and indirectly depends on the edge length required to close the cycle without imposing neither a formal definition for the cycle nor a hard threshold in edge length.

Even though observed in the example of Fig.~\ref{fig:Voronoi}, the retrieval of the $K-1$ edges of the MST is not theoretically guaranteed. To ensure it, the resulting set of edges is obtained by the union of all MST edges and those in the high probability mode leading to the non-singular symetric adjacency matrix
\begin{equation}
    \left(\boldsymbol{A}\right)_{ij} = \operatorname*{max} \left( \left(\boldsymbol{A}_{\text{MST}}\right)_{ij}, \left(\bar{\boldsymbol{A}}_{>m}\right)_{ij} \right),
\end{equation}
with $\bar{\boldsymbol{A}}_{>m}$ the thresholded average adjacency matrix $\bar{\boldsymbol{A}}$ at level $m$ such that it isolates the high probability mode only. By doing so, the proposed graph construction is an extension of the MST containing all its edges with additional ones sharing similar probabilities of appearance during all the random computations.

For the sake of illustrative comparison, we add, as the turquoise blue line in the top panel of Fig.~\ref{fig:Voronoi}, the result of the cycling topology as defined by the framework of persistent homology \cite{Edelsbrunner2002}. In particular, we use the 1D-homologically persistent skeleton of \cite[1D HoPeS,][]{Kurlin2015} built as the MST completed with edges that are creating persistent 1D homologies in the dataset. After the extraction of those critical edges, it is not trivial to obtain a persistence threshold that captures the desired cycles. Here, we rely on the bootstrap procedure that, together with the stability of the persistence diagram allows a statistically well-defined computation of a threshold \cite{Chazal2018}. The high sensitivity of this construction to outliers creates some high-scale loops leading to spurious branches corrupting the representation of the underlying pattern.
Even though these undesired cycles could be handled using the sub-level sets of the distance-to-measure function of \cite{Chazal2011, Chazal2018}, this does not allow an easy extraction of a graph on the inputted datapoints as required by our application. A pre-processing of the data by removing points in low-density areas or a careful fine-tuning of the persistence threshold are also possible but this is at odds with the proposed formalism meant to handle such datapoints.
One hint that these spurious branches come from the noisy measurements is that, when varying the randomness of the sampling, a different pattern is drawn (plotted as the orange line on the top panel of Fig.~\ref{fig:Voronoi}) while the one traced by the proposed average graph remains the same. It is also worth emphasizing that, in both realizations of the persistent skeleton, the two small-scale cycles at the center of the dataset are not captured while they are observed in our average graph prior. Since the persistence is linked with the cycle size, the incapacity in representing small cycles is caused by the too high persistence threshold imposing a lower-bound on the scales of the detected cycles.

%Figure Voronoi

\subsection{Convergence and time complexity} \label{section_convergence}

Convergence toward a local maximum and monotonic increase of the log posterior through iterations are guaranteed by the EM procedure when fixing $\forall t, \mathcal{G}^{(t)} = \mathcal{G}^{(0)}$ \citep[see][for a detailed analysis of EM convergence properties]{McLachlan1997}. When updating the topology at each iteration and using a prior based on the MST, convergence as well as monotonic increase of the regularized likelihood are still guaranteed \citep{Mao2017}, but it does not remain true for any general graph construction if not based on optimization procedures such as defined by Eq.~\eqref{MST_opt_problem}.

The E-step Equation~\eqref{Eupdate} requires the computation of the responsibilities, taking $O(NDK)$ operations to complete, while the M-step equations \eqref{Mupdate} respectively require $O(NK), O(NK), O(K^3) \text{ and } O(NKD)$ operations.
Considering $T$ iterations for the algorithm to converge and a complexity of $C_\mathcal{G}$ for the graph structure computation, algorithm \eqref{Algo:GRMM} needs $O\left(T \left[ K^3 + NKD + C_\mathcal{G} \right]\right)$ operations. $C_\mathcal{G}$ acts on $K$ and can take very different forms depending on the used topological approximation. For the MST, it takes $O(K \log K)$ operations to build the Delaunay Tessellation and $O(K^{1+D/2}\log K)$ to find the MST with Kruskal's algorithm. Naturally, the proposed average graph being obtained from $B$ random sub-samplings of the MST, it requires $O\left(B\,C_\text{MST}(K_b) + C_\text{MST}(K) \right)$ operations, with $C_\text{MST}(K)$ the complexity of getting an MST over $K$ vertices. To give a broad idea of the computational time of the algorithm, the principal graph from Fig.~\ref{fig:Voronoi} is obtained in less than three minutes on a modern laptop with our single-core Python implementation.

\subsection{Hyper-parameters}

%Figure: Illustration of the algorithm
\begin{figure*}
    \centering
    \includegraphics[width=1\linewidth]{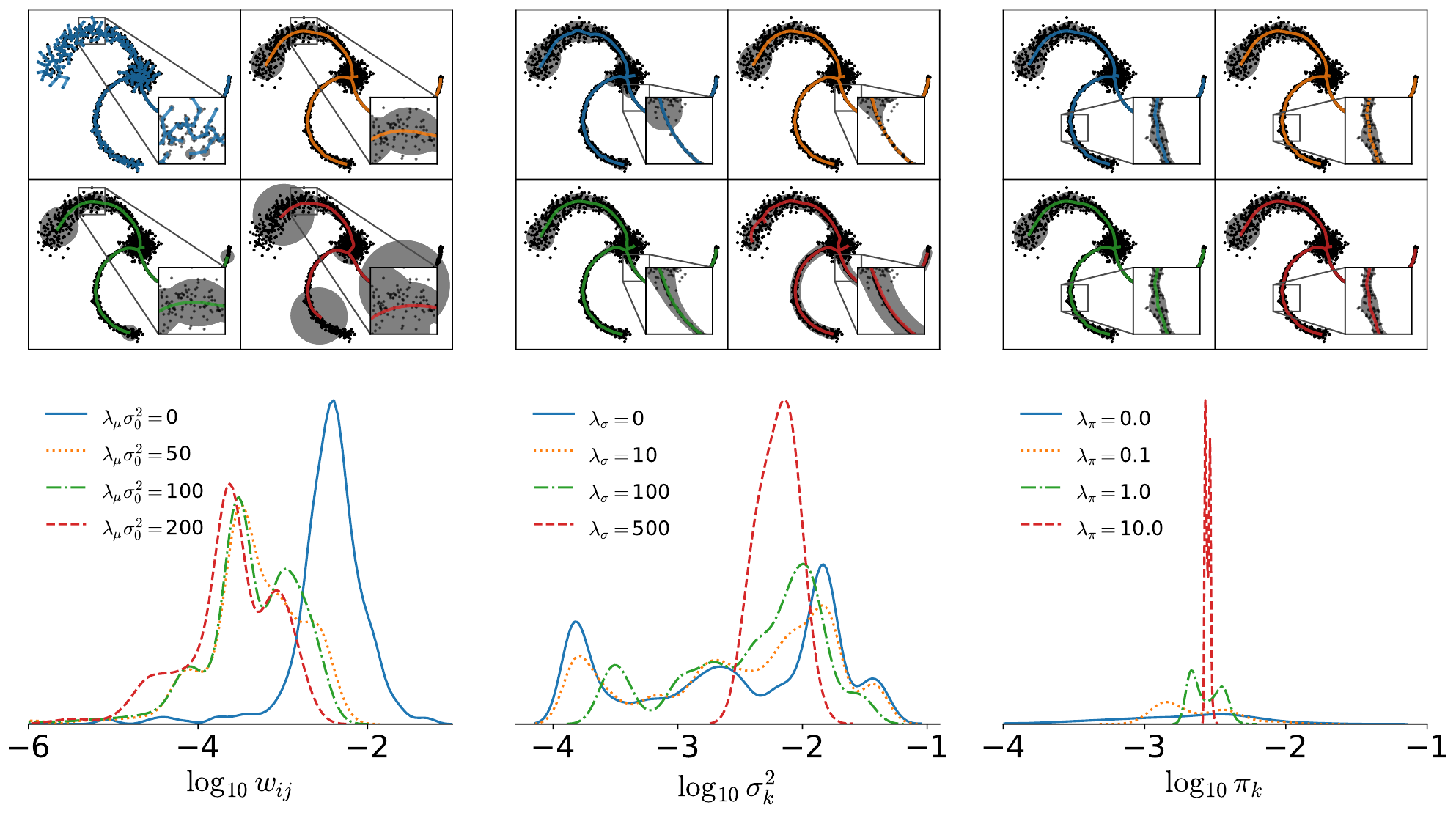}
    
    \caption{Illustration of the impact of hyper-parameters $\mathcal{Y} = \{\lambda_\mu, \lambda_\sigma, \lambda_\pi \}$ in Algorithm \eqref{Algo:GRMM} for an artificial dataset made of three branches with linearly evolving standard deviations converging to a spherical Gaussian cluster. In all cases, $K=350$ nodes are used with the same random initialization. Black dots are datapoints, colored lines are the set of edges learned by Algorithm \eqref{Algo:GRMM} and gray shaded areas correspond to the 1-$\sigma_k$ circles centered on $\boldsymbol{\mu}_k$. \textit{(top row)} From left to right, quadrants corresponds to several values of $\lambda_\mu$, $\lambda_\sigma$ and $\lambda_\pi$.
    \textit{(bottom row)} Probability distributions of the impacted parameters. From left to right: edge weights $w_{ij}=\lVert \boldsymbol{\mu}_i - \boldsymbol{\mu}_j \rVert_2$, variances of components $\sigma_k^2$, and mixing coefficients $\pi_k$.}
    
    \label{fig:Parameters}
\end{figure*}

Hyper-parameters of the full model are $K$ and $\mathcal{Y} = \left(\lambda_{\mu}, \lambda_{\sigma}, \lambda_{\pi} \right)$. $K$ denotes the number of Gaussian components used in the mixture model while $\lambda_{\mu}, \lambda_{\sigma}, \lambda_{\pi}$ are all related with shapes of prior distributions on the parameter indicated as subscripts.

$\lambda_{\mu}$ controls the force of smoothness constraint and corresponds to the precision of the prior Gaussian distribution put on edge weights. From Eq.~\eqref{prior_mu}, and emphasized by the upper left quadrant of Fig.~\ref{fig:Parameters} showing regularized graphs obtained with increasing values of $\lambda_{\mu}$ from $0$ to $200/\sigma_0^2$, we see that the higher $\lambda_{\mu}$, the more $\{\boldsymbol{\mu}_k\}_{k=1}^K$ are pulling themselves along the graph structure leading to constrain the full graph length $\sum_{ij} w_{ij}$ by shortening its extremities. The bottom left panel of Fig.~\ref{fig:Parameters} shows the final distribution of edge weights, in other words the distances between linked nodes in the graph, $w_{ij}$. We clearly observe the mode of the distribution being translated to lower values when $\lambda_{\mu}$ increases, indicating lower and lower distances between connected nodes. At a fixed value of $\lambda_\mu$, a large increase of the number of nodes $K$ can induce over-fitting. Sticking with a similar extracted pattern would hence require the increase of $\lambda_\mu$ as well. This is shown for instance when comparing the top right panel of Fig.~\ref{fig:Illustration_algo} and the top left panel of Fig.~\ref{fig:Parameters}. One way to reduce the interlink between these two parameters and the dependency of the pattern on $K$ is the growing grammar proposed in \cite{Gorban2009, Albergante2020} in which the graph is initialized with a small value of $K$ to then add some nodes iteratively using predefined rules. It also has the advantage to reduce the computational cost at early iterations.

$\lambda_{\sigma}$ acts on the shape of the inverse-Gamma prior distribution of Gaussian variances $\sigma_k^2$. Equation~\eqref{Mupdate} teaches us that, as for other hyper-parameters, when $\lambda_{\sigma} \to 0$, the prior is canceled and variances are updated through the usual EM equation. When $\lambda_{\sigma}$ increases, the shape of the inverse-Gamma law is more and more constrained leading $\sigma_k^2$ to be updated mainly through $\sigma_{\mathcal{N}_k}^2$ and, $\forall k, \sigma_k^2 \simeq \sigma_0^2$ with $\sigma_0^2$ the value used to initialize the algorithm. Eventually, a high enough value for this parameter thus leads to a fixed scale version of the algorithm, similar to those presented in \cite{Mao2017, Bonnaire2020}, as shown in the middle bottom panel of Fig.~\ref{fig:Parameters} where the distribution of $\log_{10} \sigma_k^2$ is more and more centered around $\log_{10} \sigma_0^2$.
As illustrated in the upper middle quadrant of Fig.~\ref{fig:Parameters}, when $\lambda_\sigma$ increases, the structure shows a smoothly evolving variance along the graph structure with multiple scales described at the same time. This is also highlighted by the lower panel where the variance distributions obtained from regularized graphs with small values of $\lambda_{\sigma}$ are spreading over several order of magnitudes and show three distinct modes corresponding to the characteristic scales of the three branches in the dataset. These estimates tend to be more and more biased toward $\sigma_0$ when $\lambda_\sigma$ increases.

Finally, $\lambda_{\pi}$ acts on the amplitude $\pi_k$, corresponding to the proportion of points being represented by the component $k$ and controls the force of the prior of uniformly distributed Gaussian mixtures.
This parameter has a very low impact on the overall result but can avoid singular solutions to happen in very low-density regions paved by graph nodes.
The upper right panel of Fig.~\ref{fig:Parameters} shows this small impact despite a large range of tested values, from $0$ to $10$. Although the obtained distributions of mixing weights $\pi_k$ are clearly different, as seen in the bottom right panel tending to be centered at $\left(1-\alpha\right)/K$ when $\lambda_\pi$ increases, the obtained graph structure remains unchanged, and so is the local extent measured by variances.

Even though the impact of these hyper-parameters is interpretable and that there is a wide range of values providing similar results, their quality still depend on the settings. To the best of our knowledge, there is no well-defined method to choose regularization parameters independently from user tests or external information.
In that regard, the adaption of the modified objective function from \cite{Gerber2013} in the context of principal graphs learning could be of interest to propose regularization-free models.

\subsection{Initialization}

There are four sets of parameters to initialize, namely $\{\boldsymbol{\mu}_k\}_{k=1}^K$, $\{\sigma_k\}_{k=1}^K$, $\{\pi_k\}_{k=1}^K$ and $\alpha$.
A simple and direct strategy to initialize positions of Gaussian components is to choose $\boldsymbol{\mu}^{(0)} = \boldsymbol{X}$. By doing so, we are ensured that the input point cloud distribution is well paved by centers. Note however that, from Sect. \ref{section_convergence}, when $K \simeq N$, the complexity scales with $N^3$. For large datasets, it may be interesting to first reduce the complexity by initializing the model with $K \ll N$ using for instance sub-samplings, noise reduction techniques or simple clustering methods like the K-Means or fiducial GMM algorithms.
When no prior knowledge on the local size of structures, variances can be initialized as $\forall k \in \{1, \ldots, K\}, \sigma_k^{(0)} = \sigma_0^{(0)}$. In this context, $\sigma_0^{(0)}$ can be chosen as a prior guess on the average size of structures or through rules borrowed from density estimation methods \citep{Heidenreich2013}. The $\alpha$ parameter is the evaluation of the outlier level in the dataset that should not be paved by the Gaussian components. Its value depends on the application and data at hand. In our experiments, we fix $\alpha^{(0)} = 0.10$ and then let it adjust itself during the learning. As an example, results from Fig.~\ref{fig:Parameters} were obtained by starting with this guess and then quickly converges toward a value numerically indistinguishable from $0$. Finally, mixing coefficients are initially assumed to be uniformly distributed and $\forall k \in \{1, \ldots, K\}, \pi_k^{(0)} = \left(1-\alpha^{(0)}\right)/K$.

One of the known major drawback of the EM algorithm is its dependency on parameter initialization due to its tendency to be easily trapped in local maxima of complexly shaped posterior distributions. To counter that problem, solutions based on simulated annealing (SA) were introduced \citep{Ueda1998}. \cite{Bonnaire2021} shows how SA can be used jointly with regularization of GMMs to learn a heteroscedastic graph representation independently of the initialization. However, this solution comes with a sizable increase of the computational cost and, in all our experiments, we did not find it necessary to obtain satisfactory results.

%==========================================================================
\section{Applications} \label{section_applications}

In what follows, we focus on two applications where the newly introduced features, namely the handling of background noise, the adaptive scale and the cycling topology, are of interest.
Figures \ref{fig:Illustration_algo} to \ref{fig:Parameters} already introduced example of applications for general manifold denoising purposes in case of heteroscedastic sampling. These approaches can generally be used as pre-processing for clustering or to reduce the dimensionality of the data. To illustrate the use of the GRMM, we tackle two problems in very different fields, namely the automatic detection of filaments in the galaxy distribution and the identification of road networks from noisy GPS measurements. Even though the two applications are of low dimensionality to showcase results of the algorithm, the GRMM can also be used to identify 1D structures in higher-dimensional input spaces. This comes with an increase of the time complexity but also with the need of more datapoints to accurately trace the pattern.

\subsection{Filamentary structure detection in galaxy distribution} \label{section_application_cosmo}

\begin{figure}
    \centering
    \includegraphics[width=1\linewidth]{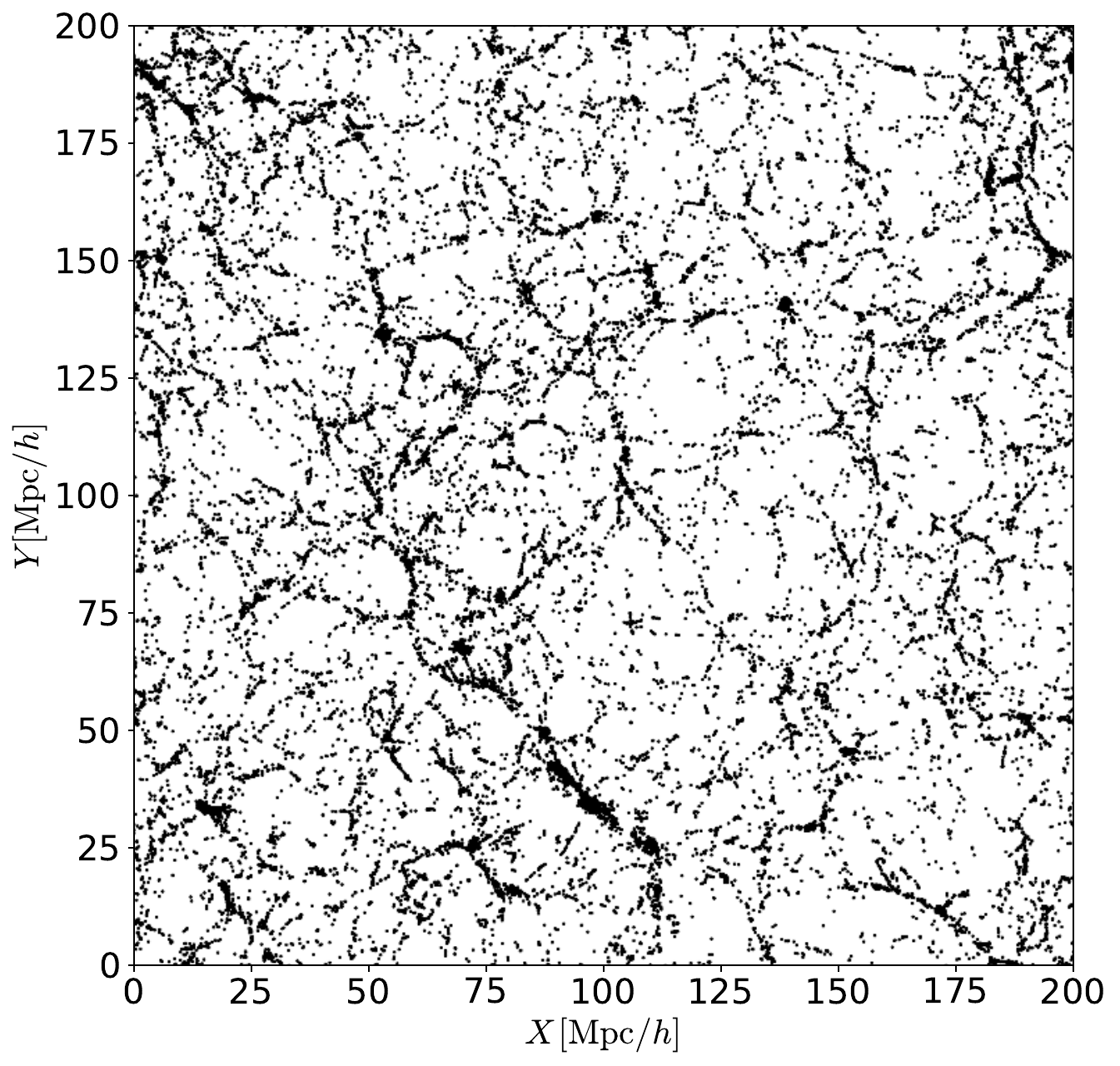}
    
    \caption{A 2D distribution of $N=31500$ galaxies obtained from the IllustrisTNG simulation over a thin slice of $0.5$ Mpc/$h$.}
    
    \label{fig_TNG_dp}
\end{figure}

\begin{figure*}
    \centering
    \includegraphics[width=1\linewidth]{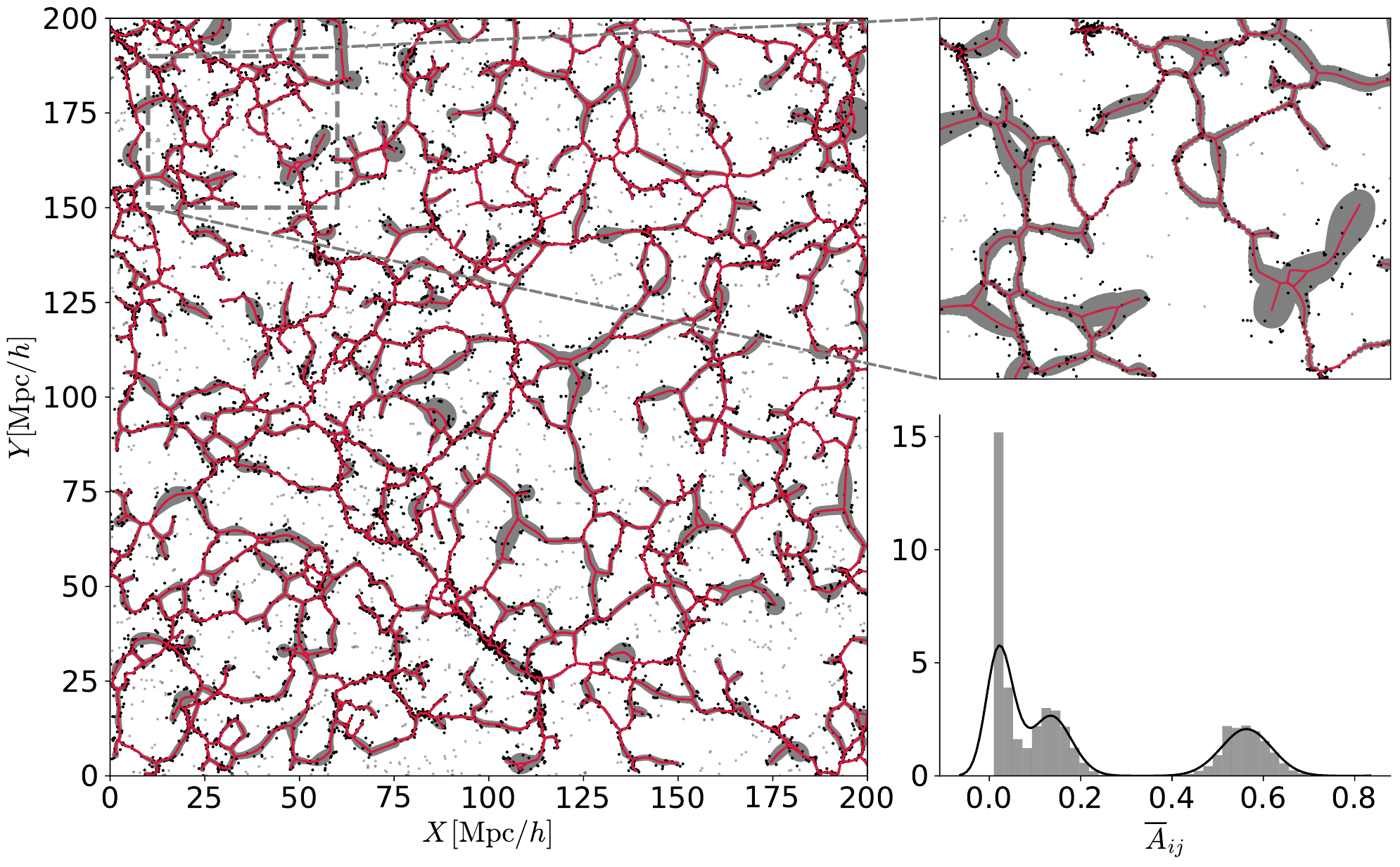}
    
    \caption{\textit{(left)} Black (resp. gray) points are those with $\sum_{k=1}^K p_{ik} > p_i^\text{bkg}$ (resp. $<$). The red skeleton corresponds to the regularized graph obtained from Algorithm \eqref{Algo:GRMM} with the average graph prior and settings $K = 13390, \sigma_0^2 = 1, \mathcal{Y} = \{10, 5, 1\}$. Gray shaded areas are showing the learned 1-$\sigma_k$ circles around graph nodes. \textit{(top right)} Zoom over a $60$ Mpc/$h$ region. Color code is the same as for the left panel. \textit{(bottom right)} Probability distribution function of $\left(\bar{\boldsymbol{A}}\right)_{ij}$ for minimum spanning trees obtained from $B=500$ random sub-samplings of Gaussian centers with $K_b/K= 0.75$.}
    
    \label{fig_TNG}
\end{figure*}

In observational cosmology, a recurrent problem is the identification and extraction of cosmic structures from the spatial distribution of galaxies in the universe. This distribution, far from being uniform shows multiple modes and vast empty areas. Among these structures is the striking filamentary pattern linking together regions of high density. One key issue in this problem is to efficiently detect these filaments from a point cloud distribution of observed galaxies to further assess astrophysical hypotheses and improve our understanding of phenomena driving the structure formation of the universe. Here, we rely on a large-scale simulation, namely the IllustrisTNG\footnote{\url{https://www.tng-project.org/}} simulation \citep{Nelson2019}, where massive particles are evolved forward in time from the primordial universe to present time through gravity equations giving birth to a structured pattern visible on the thin slice of $0.5$ Mpc/$h$ width of Fig.~\ref{fig_TNG_dp}. Such data are known to require particularly elaborated methods to be processed with pattern exhibiting structures of various densities and sizes, from large arteries to tenuous tubular filaments, embedded in a noisy background of galaxies standing in quasi-empty areas.
In this cosmological context, the MST definition has been applied in the past \citep{Barrow1985} and provides a parameter-free network linking all galaxies together. As we will see, the proposed extensions bring a smooth behavior for branches of the graph (filaments), the description of their local extent, and the representation of cycles in the network.

Figure~\ref{fig_TNG} shows the result of the GRMM Algorithm \eqref{Algo:GRMM} applied on the 2D slice of Fig.~\ref{fig_TNG_dp}, together with the probability distribution of edge frequencies to build the average graph presented in Sect. \ref{subsection_averageGraph}. Although the dataset presents complex patterns and long-range features, we still observe two populations and, among edges with high probabilities, we add to the MST 132 edges that are, by construction, creating cycles in the datasets. The visual inspection of the obtained cycling graph on the zoomed top-right panel of Fig.~\ref{fig_TNG} shows that the approach is closing the graph in relevant places, resulting in a much richer structure that if we had used the MST only. The gray shaded regions illustrate the local size of the filamentary pattern learned by the algorithm. Variances $\{\sigma_k^2\}$ are following the radial extent of datapoints but also show a coherent smooth evolution.

The obtained principal modeling the cosmological filamentary pattern leads to a smooth estimate of the ridge passing through galaxies which is not always the case in other extractors \citep{Libeskind2017}.
Additionally, our algorithm provides the connection between the Gaussian components through the graph structure allowing to define individual filaments which is not naturally allowed, for instance, for projected points of SCMS-based methods \citep{Chen2014}. 
To a further extent, the resulting graph also allows the extraction of geometrical information on the structure, such as the local width or length of the pattern and an estimate of the local density, that are of high interest for astrophysical studies.

Finally, the probabilistic setup of the mixture model can provide a denoised version of the pattern. By keeping only the set of datapoints such that $\{\boldsymbol{x}_i \, | \, \sum_{k=1}^K p_{ik} > p_{i}^\text{bkg}\}$, i.e. those with a probability of being generated by the Gaussian components paving the pattern higher than being generated by the background noise, we remove datapoints in which there is no identified spatial structure. Such datapoints are shown in gray in Fig.~\ref{fig_TNG} leading to a cleaned version where remains only galaxies inside the filamentary network plotted in black in the figure.

\subsection{Road network extraction from noisy GPS measurements}

\begin{figure}
    \centering
    \subfigure{\includegraphics[width=0.78\linewidth]{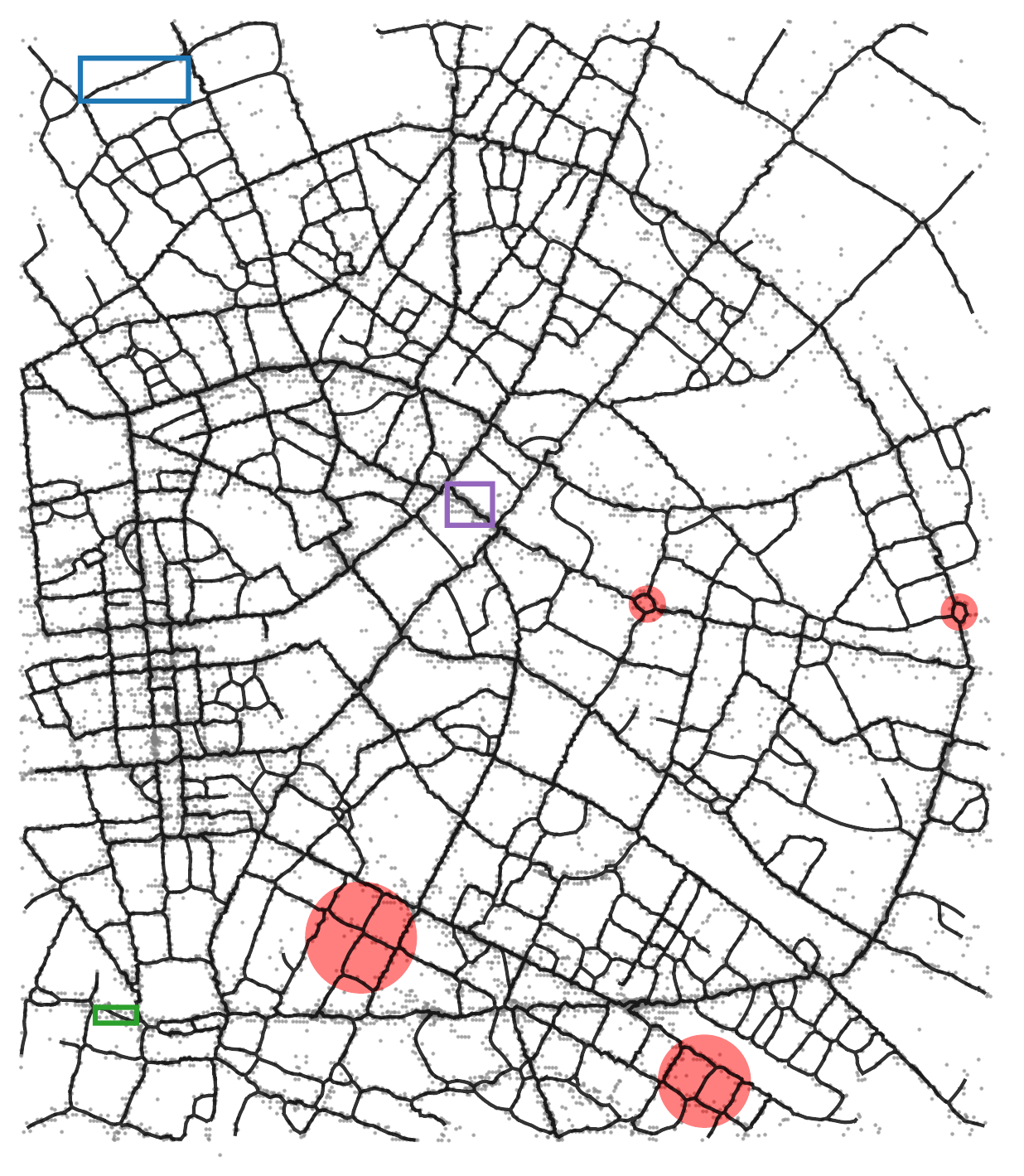}}
    \subfigure{\includegraphics[width=0.78\linewidth]{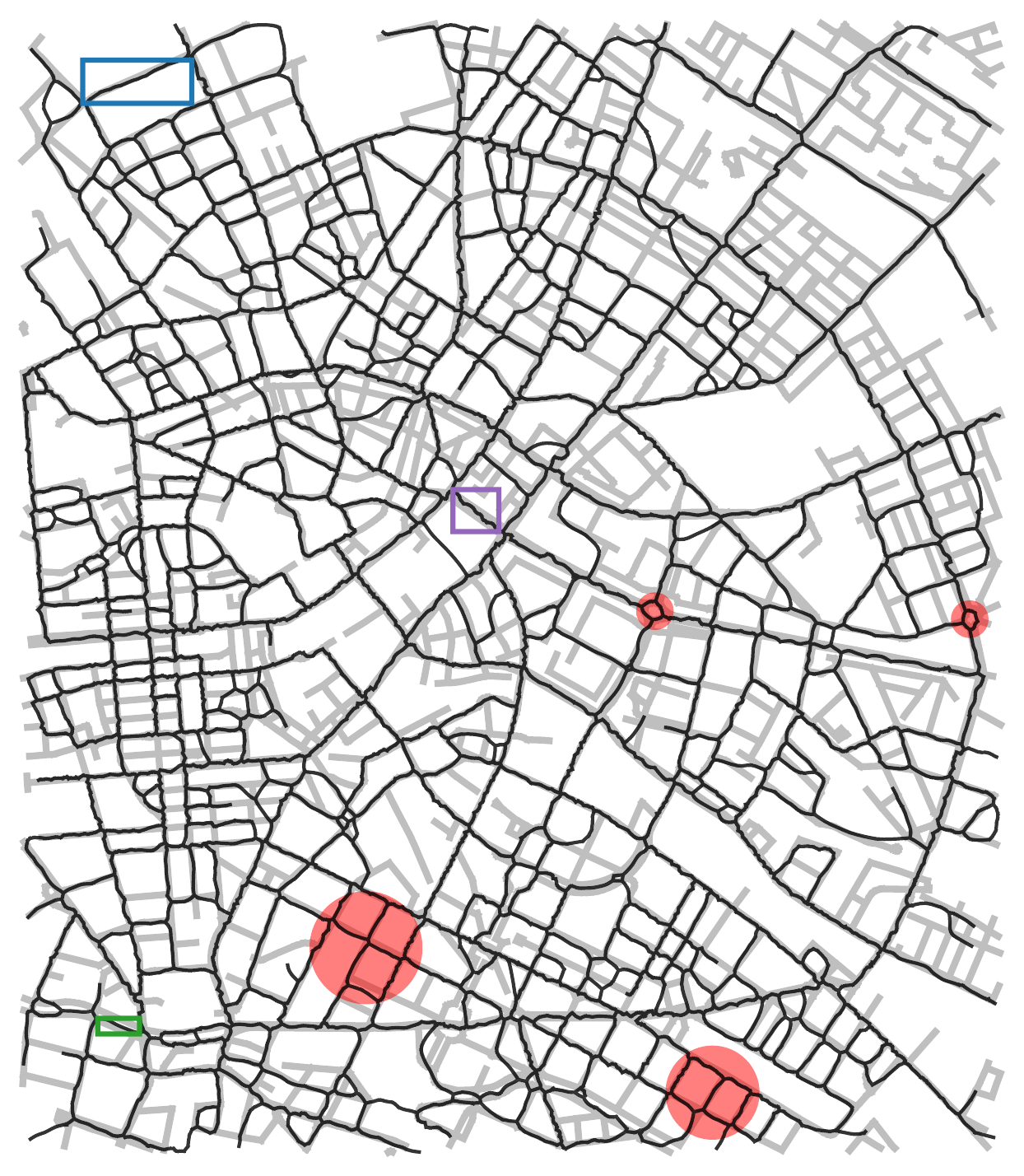}}
    \subfigure{\includegraphics[width=0.79\linewidth]{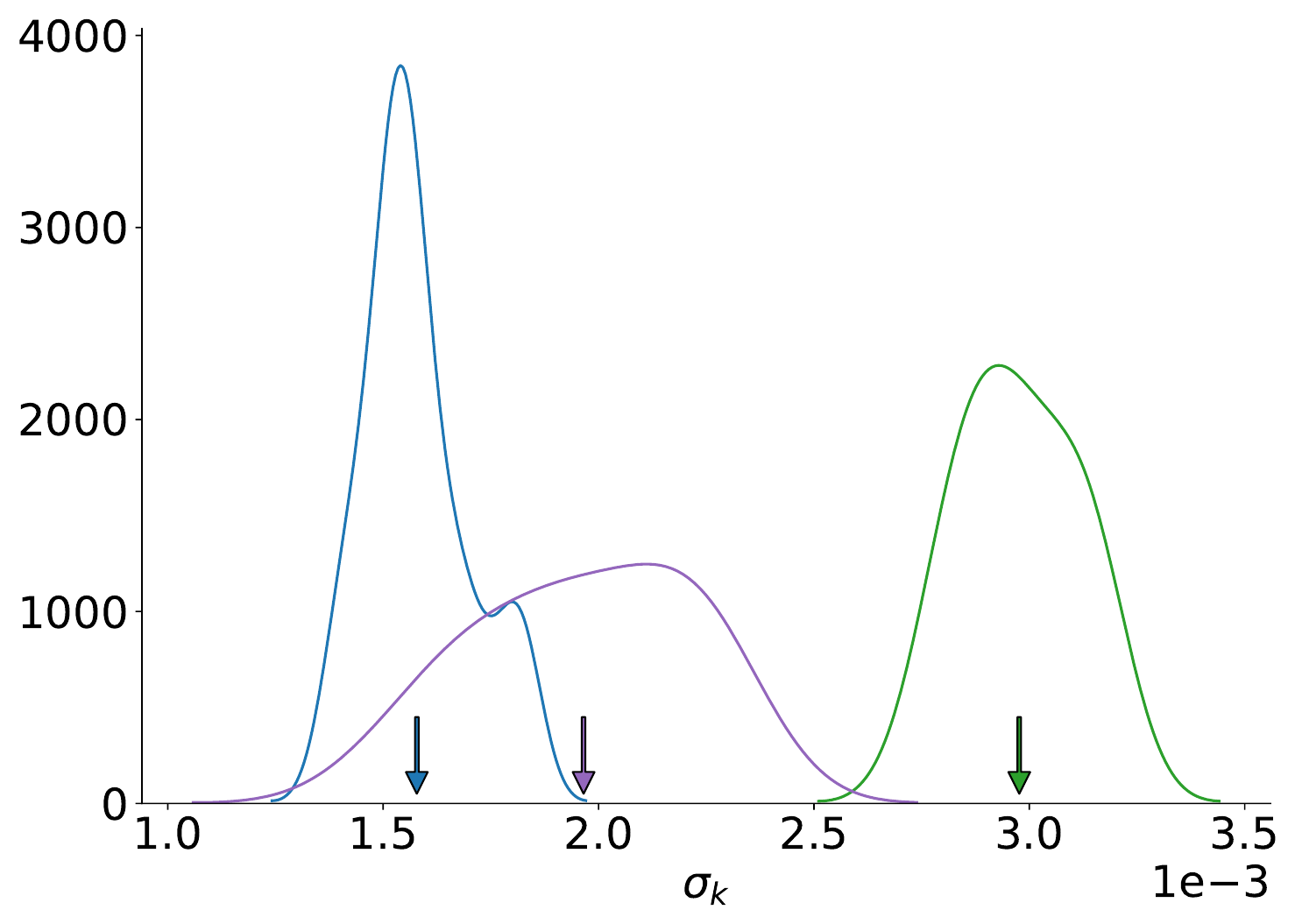}}
    
    \caption{\textit{(top)} Black lines are edges of the regularized graph of Algorithm \eqref{Algo:GRMM} overplotted on the raw datapoints. The graph was obtained using the average graph topology with $K=7300$ and $\sigma_0 = 0.003$, $\mathcal{Y} = (10/\sigma_0^2, 5, 1)$. Red circles highlight features that could not be caught by a non-cycling topology.
    \textit{(bottom)} Probability distributions of variances for nodes in the corresponding rectangular colored regions of the top panel.}
    \label{fig:Berlin}
\end{figure}

Producing street maps from sampled positions of traveling vehicles is one of the objectives of map reconstruction. In this case, the set of measurements is obtained from sampled GPS positions of taxis in Berlin\footnote{\url{http://mapconstruction.org/}}, a classical benchmark dataset of road network reconstruction \citep{Ahmed2015}. These data are especially challenging due to their high level of noise and outliers but also because of the uneven sampling of several trajectories. Main roads are indeed traced by hundreds of trajectories, while some low-level ones are traced only by a few.

When applying the GRMM Algorithm \eqref{Algo:GRMM} with the average graph prior on the Berlin dataset, we obtain the graph from the top panel of Fig.~\ref{fig:Berlin} with no pre-processing of the dataset. We see that the principal graph could use some topological post-processing refinements to make it look smoother, removing waving branches and closing some of the remaining dangling ones. For more details about particular processing of inputs and outputs in the context of graph structure learning for road networks, see \citep{Huang2018}. The obtained graph passes through the entire dataset, correctly paving regions with both high and low densities. The proposed algorithm also recovers a large variety of cycle sizes, from very small ones, as for instance in the red shaded region with intersections or roundabouts, but also larger ones, as in the top left corner.
When comparing it with the ground truth map obtained from OpenStreetMap\footnote{\url{https://www.openstreetmap.org/}} in the middle panel, we clearly distinguish that some roads are not well-traced, with several outliers due to the poor quality of sampled data, as it is the case in the top right corner. The GRMM however succeeds in proposing a robust version that does not take into account all such datapoints.

The proposed method additionally offers the possibility to estimate the local variance around the inferred principal graph. In the present dataset for instance, it is possible to investigate further small portions of roads where the estimated $\sigma_k$ is high or has a broad distribution to spot roads that are either wider or noisier than others. A quantitative comparison to other existing methods is beyond the scope of the present paper as this application only serves an illustrative purpose. The bottom panel of Fig.~\ref{fig:Berlin} reports the distribution of variances for nodes standing in rectangular regions of the top panel. When using variances of Gaussian clusters as a proxy of the road width, we conclude that main roads with multiple lanes like those in the green or purple rectangles have larger sampling standard deviation, by up to a factor of two, than the low-level road in the blue rectangle.
These estimates of road widths however, depend on the quality of the sampling and some roads can appear wider than they actually are because of spurious points artificially increasing the estimate locally or at the extremities of some branches of the graph. This can be seen when inspecting the variance of the $\sigma_k$ distributions where the purple one is much larger than the others because of the local noise in the sampling.
Finally, the proposed cycling graph captures correctly some topological features of the road network, such as the most prominent roundabouts or road intersections highlighted in shaded reddish areas on the two top panels of Fig.~\ref{fig:Berlin}. The representation of such features would not be permitted by tree-based topologies.

%==========================================================================
\section{Conclusions} 
We present an approach for the learning of principal graphs from point cloud datasets based on mixture models and graph regularization. The proposed formalism provides a robustness to outliers of the pattern and can represent structures of various sizes that are estimated iteratively during the learning. We showed that the added features of heteroscedastic learning and outliers embedding come with only little modifications and additional time complexity of already existing procedures and compared it to state-of-the-art methods.
Beyond that, one of our contributions is the introduction of a stochastic procedure to infer the principal graph, where the construction of many MSTs on sub-sampled datasets generate a graph with cycles of several scales, features exhibited in many real-world datasets. Not only the overall procedure is simple to implement but it also relies on fast and well-established algorithms, namely EM for the mixture model and Kruskal for the MSTs, making it particularly appealing from a computational point of view. The whole optimization scheme is meant to converge toward a local maximum of the log-posterior for any graph prior. The method also shows a stable behavior for a large range of model's hyper-parameters and initialization, giving freedom to the user to retrieve the underlying patterns without requiring a cautious fine-tuning.
Finally, we illustrated the importance of these newly introduced characteristics in two applications using point-cloud distributions as input, namely the reconstruction of road networks from vehicles position and the detection of the filamentary pattern of the cosmic web. In both applications, even though the data show a high level of complexity with different local sampling densities, loops of various sizes, noises, features of different scales and that no pre-processing were employed to remove spurious measurements, the obtained cycling networks visually trace the expected underlying patterns.
We also leave for further investigations the adaption of the learning of locally higher dimensional embeddings that could be done using different graph priors as exploited in other manifold learning frameworks \citep{Belkin2003, Gorban2005}.

\section*{Acknowledgments}
T.B. thanks Dr. Victor Bonjean for fruitful discussions. This research has been supported by the funding for the ByoPiC project from the European Research Council (ERC) under the European Union’s Horizon 2020 research and innovation program grant number ERC-2015-AdG 695561. A.D. was supported by the Comunidad de Madrid and the Complutense University of Madrid (Spain) through the Atracción de Talento program (Ref. 2019-T1/TIC-13298).

\bibliographystyle{IEEEtran}
\bibliography{MM_theory}

\end{document}